\documentclass{article} 
\usepackage{iclr2026_conference,times}

\usepackage{amsmath,amsfonts,bm}

\def\eqref#1{equation~\ref{#1}}

\def\1{\bm{1}}

\DeclareMathAlphabet{\mathsfit}{\encodingdefault}{\sfdefault}{m}{sl}
\SetMathAlphabet{\mathsfit}{bold}{\encodingdefault}{\sfdefault}{bx}{n}

\usepackage{caption}

\usepackage{url}
\usepackage[utf8]{inputenc} 
\usepackage[T1]{fontenc}    
\usepackage{hyperref}       
\usepackage{url}            
\usepackage{booktabs}
\usepackage{amsfonts}       
\usepackage{nicefrac}       
\usepackage{microtype}      
\usepackage{xcolor}         
\usepackage{amsmath}
\usepackage{xspace}
\usepackage{amssymb}

\usepackage{tikz}
\usepackage{graphicx}

\usepackage{booktabs}       
\usepackage{graphicx}
\usepackage{multirow}
\usepackage{multicol}
\usepackage{color}
\usepackage{colortbl}
\definecolor{Gray}{gray}{0.86}
\usepackage[most]{tcolorbox}
\usepackage{tabularx}

\usepackage{algorithm}
\usepackage{algpseudocode}

\usepackage{ulem}

\usepackage{amssymb}
\usepackage{bbding}
\usepackage{pifont}
\usepackage{wasysym}
\usepackage{amssymb}
\newcommand{\grpo}{MaskGRPO\xspace}
\definecolor{do}{rgb}{0.8, 0.3, 0.0}
\definecolor{db}{rgb}{0.4, 0.3, 0.8}

\title{Consolidating Reinforcement Learning for \\Multimodal Discrete Diffusion Models}

\vspace{-0.1cm}
\author{Tianren Ma, \ Mu Zhang, \ Yibing Wang, \ Qixiang Ye\\
University of Chinese Academy of Sciences\\
{\tt\small matianren18@mails.ucas.ac.cn;qxye@ucas.ac.cn}
}

\iclrfinalcopy 
\begin{document}

\vspace{-0.4cm}
\maketitle





\vspace{-0.5cm}

\begin{abstract}

Optimizing discrete diffusion model (DDM) with rewards remains a challenge—the non-autoregressive paradigm makes importance sampling intractable and rollout complex, puzzling reinforcement learning methods such as Group Relative Policy Optimization (GRPO).
In this study, we introduce \grpo, the first viable approach to enable scalable multimodal reinforcement learning in discrete diffusion with effective importance sampling and modality-specific adaptations.
To this end, we first clarify the theoretical foundation for DDMs, which facilitates building an importance estimator that captures valuable token fluctuation for gradient updates.
We then delicately tailored the rollout method for visual sequences, which yields diverse completions and reliable optimization gradients.
%
%
Upon math reasoning, coding, and visual generation benchmarks, \grpo brings more stable and efficient updates, leading to stronger reasoning performance and better generation quality. 
This study establishes \grpo as a systematic policy optimization approach and the first practical way for discretized visual diffusion. Our code is available at \url{https://github.com/martian422/MaskGRPO}.
%
\end{abstract}
\begin{figure}[!ht]
 \centering
\includegraphics[width=\linewidth]{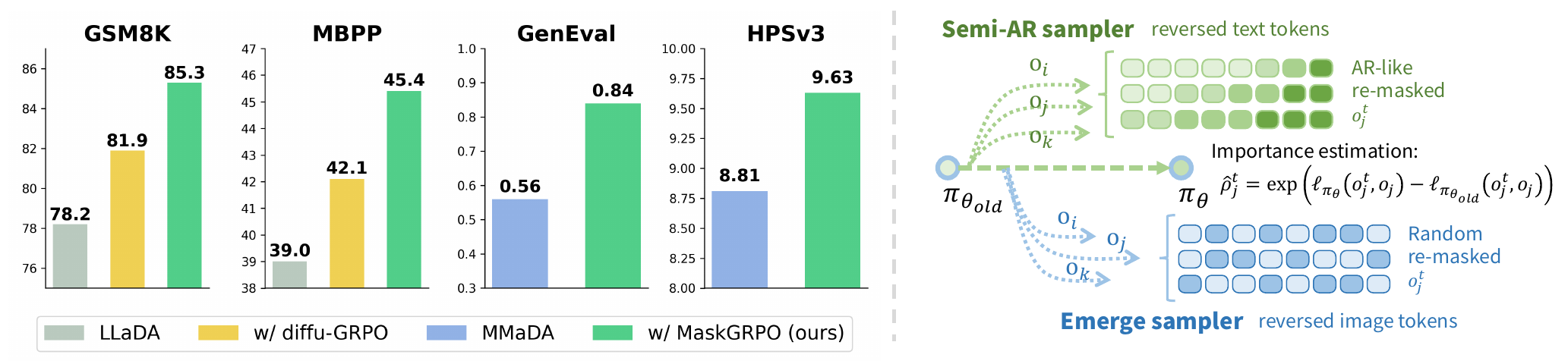}
\caption{Left: \textbf{\grpo} consistently improves the base model with significant RL income across text and image generation tasks. Right: an intuitive demonstration of our method, integrated with modality-specific innovations on importance estimation and sampling methods.}
\label{fig:intro}
\end{figure}
\section{Introduction}
Recent progress of post-training generative models has been driven by the advances of optimization algorithms, architectural design, and large-scale reward-based learning~\cite{rafailovDirectPreferenceOptimization, wangUnifiedRewardModel2025,liuFlowGRPOTrainingFlow2025}.
Among these, Group Relative Policy Optimization (GRPO)~\cite{shaoDeepSeekMathPushingLimits2024} has emerged as a powerful and scalable paradigm, improving reasoning performance of large language models and enhancing preference alignment of visual generative models.
However, extending such policy optimization to discrete diffusion models (DDMs) remains a challenge.

Unlike autoregressive models that decode sequentially, discrete diffusion generates tokens in parallel at arbitrary positions~\cite{sahooSimpleEffectiveMasked2024}.
This parallelism complicates both \textit{rollout generation}, where stochastic yet coherent samples are required~\cite{liuFlowGRPOTrainingFlow2025} for exploration, and \textit{importance estimation}, which is crucial for optimization~\cite{schulmanProximalPolicyOptimization2017}.
%
Existing approaches offer only partial solutions: semi-autoregressive samplers~\cite{arriolaBlockDiffusionInterpolating2025,nieLargeLanguageDiffusion2025a} mitigate inference issues for text, while low-confidence re-masking for images~\cite{changMaskGITMaskedGenerative2022} lack stochastic flexibility for robust group comparisons.
Similarly, early attempts~\cite{zhaoD1ScalingReasoning2025,tangWd1WeightedPolicy2025a} at importance estimation relied on masking heuristics that violated conditioning assumptions.
Monte Carlo–based estimators~\cite{zhuLLaDA15VarianceReduced2025,yangMMaDAMultimodalLarge2025} improve faithfulness but remain computationally expensive.

In this study, we propose \textbf{\grpo}, a consolidated extension of GRPO to multimodal discrete diffusion (shown in Fig.~\ref{fig:intro}), built upon modality-specific innovations in both sampling and importance estimation.
\grpo is guided by the distinct structural properties of language and vision:

\textbf{Language.} While training native discrete diffusion models depart from the autoregressive\footnote{In this paper, we use \textit{autoregressive} in its conventional sense of causally ordered next-token prediction.} paradigm, their prediction on natural language still exhibits a degree of ``ARness"~\cite{gongDiffuCoderUnderstandingImproving2025}: tokens closer to observed context are predicted with higher certainty, and rollouts diverse as length extends. 
Leveraging this property, we introduce a fading-out masking estimator, which progressively increases the masking rate toward later tokens with well-controlled randomness.
This concentrates estimation on high-uncertainty regions, towards a more efficient and empirically reliable objective.

\textbf{Vision.} Images lack a sequential structure and exhibit strong global token correlations~\cite{chanAnalyzingLanguageVisual2024}.
We argue that effective likelihood estimation requires highly truncated mask rates to capture informative variation. Furthermore, we propose a sampler that relaxes rigid scheduling constraints in existing methods via probabilistic decoding.
By encouraging diverse yet high-quality rollouts, our sampler better aligns with the GRPO principle of exploiting group-wise relative advantages.

Through these contributions, we build the first systematic GRPO approach for multimodal discrete diffusion.
Beyond empirical results in mathematical reasoning and coding that almost double the income from RL, our method also demonstrates significant improvement on text-image alignment, and visual fidelity. Building upon a clarified foundation for DDMs, our analysis highlights that, policy optimization in discrete diffusion is only effective when samplers and estimators are designed in a modality-aware fashion.
This establishes a new foundation for reward-based learning in DDMs and points toward a more general theory of preference-driven optimization across modalities.

\section{Preliminaries}

\subsection{Discrete Diffusion Model}
DDM defines a forward process over discrete variables by gradually corrupting tokens to absorbing state $\mathbf{m}$ through a continuous-time Markov process.
We denote the clean data as $x_{t=0}$ ($x_0$ for short), and noise it gradually as $t\rightarrow 1$.
Let \( \alpha_t \) be the noise scheduler (a monotonically decreasing survival function that satisfies \(\alpha_0=1, \alpha_1=0\) ), the corrupted data distribution at time $t$ is determined as 
\begin{equation}\label{denote}
    x_t \sim q(x_t|x_0, t), q(x_t|x_0, t) = \text{Cat}(x_t; \alpha_t x_0 + (1-\alpha_t)\mathbf{m})
\end{equation}
Let $\delta(x_{(t,i)},\textbf{m})$ be the indicator function that is only activated if the $i$-th position of $x_t$ is $\mathbf{m}$.
For a linear scheduler, the objective is derived as the evidence lower bound (ELBO) of $\log \pi_\theta(x_0|x_t)$:
\begin{equation}\label{ddmloss}
    \mathcal{L}_{\text{DDM}} = -\mathbb{E}_{t,\ x_0,\ x_t} [\frac{1}{t}\sum_{i=1}^L\delta(x_{(t,i)},\mathbf{m})\log \pi_\theta(x_{(0,i)}|x_t)]=-\mathbb{E}_{t,\ x_0,\ x_t} [\ell_{\pi_\theta}(x_t,x_0)].
\end{equation}
We denote the loss term as $\ell_{\pi_\theta}(x_t,x_0)$ for later usage.
For conditional generation where a prompt $\mathbf{c}$ is given, we write $\ell_{\pi_\theta}(x_t,x_0|\mathbf{c})$ for simplicity.
Following MDLM's deduction~\cite{sahooSimpleEffectiveMasked2024}, assume that the network can reconstruct $x_0$ perfectly, we use $\pi_\theta(x_t)$ to approximate this denoising process, and get the sampling rule as
\begin{equation}\label{reverse}
    p_\theta(x_s|x_t)= 
    \begin{cases}
    1, & \text{if } x_s = x_t, \ x_t \ne \mathbf{m} ,\\
    \frac{1-\alpha_s}{1-\alpha_t}, & \text{if } x_s =\mathbf{m}, \ x_t =\mathbf{m},\\
    \frac{\alpha_s-\alpha_t}{1-\alpha_t}\pi_\theta(x_t), & \text{if } x_s \ne \mathbf{m}, \ x_t = \mathbf{m},  \\
    0, & \text{otherwise.} \\
    \end{cases}
\end{equation}

\subsection{GRPO for Autoregressive Model}
GRPO~\cite{shaoDeepSeekMathPushingLimits2024} introduces two key innovations to simplify online-RL approaches like PPO~\cite{schulmanProximalPolicyOptimization2017}. First, GRPO eliminates the explicit modeling of value function and instead computes advantages in a group-relative manner. 
This design enables GRPO focusing on relative performance within a local context, reducing sensitivity to absolute reward scales. Second, GRPO extends the clipped objective by incorporating an explicit KL divergence penalty term between the current policy $\pi_\theta$ and a reference policy $\pi_{\text{ref}}$. 

Formally, for each question $\mathbf{c}\sim \mathcal{D}$, GRPO samples a group of $G$ responses (also addressed as rollouts) $\{o_1, o_2, \dots, o_G\}$ from the old policy $\pi_{\theta_{\text{old}}}$. For rollout $o_i$, the reward system gives a action value $r_i$, and the relative advantage of it is normalized as
\begin{equation}\label{adv}
    A_i = \frac{r_i - \text{mean}(\{r_j\}_{j=1}^G)}{\text{std}(\{r_j\}_{j=1}^G)}.
\end{equation}
 For position $1 \leq k \leq |o_i|$ , the token-level importance is calculated as 
\begin{equation}\label{importance}
\rho_i^k = \frac{\pi_\theta(o_i^k|\mathbf{c}, o_i^{<k})}{\pi_{\theta_{\text{old}}}(o_i^k|\mathbf{c}, o_i^{<k})}=\exp\big(\log\pi_\theta(o_i^k|\mathbf{c}, o_i^{<k})-\log\pi_{\theta_{\text{old}}}(o_i^k|\mathbf{c}, o_i^{<k})\big).
\end{equation}
With $\epsilon$ controlling the clip range, the reward component is defined as
\begin{equation}
R(\theta,\mathbf{c}) =  
    \frac{1}{G} \sum_{i=1}^G \frac{1}{|o_i|} \sum_{k=1}^{|o_i|}
    \min \Big( \rho_i^k A_i, \, 
    \text{clip}(\rho_i^k, 1-\epsilon, 1+\epsilon) A_i \Big),
\end{equation}
Finally, the GRPO objective is expressed as a reward-penalty tradeoff, as
\begin{equation}\label{grpo}
\max_{\theta} \,\mathbb{E}_{\mathbf{c}\sim\mathcal{D}, o_{1:G} \sim \pi_\theta(\cdot|\mathbf{c})}\Big[R(\theta, \mathbf{c}) - \beta 
 \mathbb{D}_{\text{KL}} \big[ \pi_\theta(\cdot|\mathbf{c}) \,\|\, \pi_{\text{ref}}(\cdot|\mathbf{c}) \big] \Big],
\end{equation}
where $\beta$ regulates the strength of the KL regularization. 

\subsection{Accommodating GRPO for Discrete Diffusion Model}
We accommodate GRPO's objective for DDM in this section. To avoid confusion and align with settings in Eq.~\ref{grpo}, we move the timestep notation to the top-right corner of the variable. Accordingly, let $o^t \sim q(o^t|o,t)$ denote the corrupted (reversed) response $o$ with strength $t$.
Rolling back sequentially on AR model's response can be regarded as reversing on the timeline of the response. In other words, the first $k$ tokens, $o^{\leq k}$, of AR model's response can be regarded as a re-masked $o$ with $t=\frac{|o_i|-k}{|o_i|}$. Therefore, for each completion $o$ from a DDM, we can calculate the sub-sequence level importance $\rho^t$ by gradually reversing it. 

Recalling $\ell_{\pi_\theta}$ from Eq.~\ref{ddmloss}, for a small interval $\delta t$, let $\dot{o}^t = o^t - o^{t+\delta t}$ denote the tokens that are unmasked from timestep $t+\delta t$ to $t$.
We propose that the differentiation on DDM's intractable log-likelihood can be approximated (refer to the Appendix~\ref{sec:proof} for details) using
\begin{equation}\label{approx}
\log\pi_{1}(\dot{o}^{t}|\mathbf{c},o^{t+\delta t})-\log\pi_{2}(\dot{o}^{t}|\mathbf{c},o^{t+\delta t})\approx \ell_{\pi_{1}}(o^{t}, t,o|\mathbf{c})-\ell_{\pi_{2}}(o^{t}, t,o|\mathbf{c})
\end{equation}
The above expression indicates that, to evaluate the fluctuation of likelihood for newly unmasked tokens in $o^t$, we can utilize the difference of model's prediction for the full sequence at time $t$. Hence, we derive the calculable importance estimation and KL divergence as
\begin{equation}\label{rho-ddm}
\hat{\rho}_i^t =\exp\big(\ell_{\pi_\theta}(o_i^t,o_i|\mathbf{c})-\ell_{\pi_{\theta_{\text{old}}}}(o_i^t,o_i|\mathbf{c})\big).
\end{equation}
\begin{equation}\label{kl-ddm}
\hat{\mathbb{D}}_{\text{KL}}^{i,t} =\exp\big(\ell_{\pi_{\theta_{\text{ref}}}}(o_i^t,o_i|\mathbf{c})-\ell_{\pi_\theta}(o_i^t,o_i|\mathbf{c})\big)-\big(\ell_{\pi_{\theta_{\text{ref}}}}(o_i^t,o_i|\mathbf{c})-\ell_{\pi_\theta}(o_i^t,o_i|\mathbf{c})\big)-1.
\end{equation}
We temporarily skip the clip operation for simplicity, and accommodate Eq.~\ref{grpo} as
\begin{equation}\label{grpo-ddm}
\max_{\theta} \, \mathbb{E}_{\mathbf{c}\sim\mathcal{D}, o_{1:G} \sim \pi_\theta(\cdot|\mathbf{c})}\Big[\frac{1}{G} \sum_{i=1}^G \frac{A_i}{|o_i|} \sum_{j=1}^{\mu}
    (\hat{\rho}_i^{t_j}-\beta \hat{\mathbb{D}}_{\text{KL}}^{i,t_j}) \Big], \,\, \ \ t_j=j/\mu
\end{equation}
Upon this foundation, we revisit the prior endeavors on DDM optimization: \textit{diffu}-GRPO~\cite{zhaoD1ScalingReasoning2025} applies masks to prompts and extract likelihood on the entirely masked completions $o^{t=1}$.
Following LLaDA-1.5~\cite{zhuLLaDA15VarianceReduced2025}, UniGRPO~\cite{yangMMaDAMultimodalLarge2025} iteratively masks varying ratio of completions.
While these strategies provide gradient signals, they either disrupt the conditional dependency or pose high budget for Monte Carlo style estimations.
In summary, current inefficiency of likelihood estimation ties DDM to limited settings and obscures its potential in broader contexts, especially in reasoning or visual generation that may involve thousands of tokens per sample.

\begin{figure}[htbp]
    \centering
    \includegraphics[width=\linewidth]{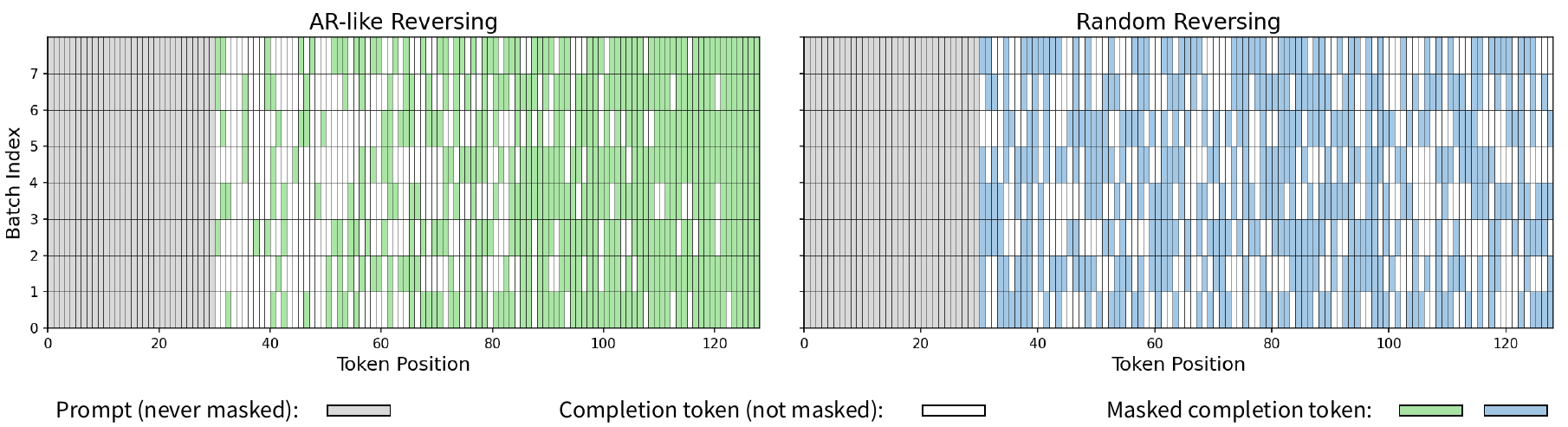}
    \caption{\textbf{A demonstration of reversing (re-mask) methods.} We set mask raio $r=0.6$. Random reversing (right) applies masks to all the tokens with equal probability, while AR-like reversing (left) adapts a fading-out strategy. See Appendix~\ref{append-samples} for complete showcases.}
    \label{fig:remask}
\end{figure}
\section{MaskGRPO}

\subsection{Importance Estimation}
Importance sampling is central to the GRPO objective, where it functions as an effective "reweighting" of the rewards, aiming to align with the current policy’s distribution with reduced bias. In practice, the importance is calculated by the differentiate of predictions as in Eq.~\ref{approx}. Inspired by recent studies~\cite{wang8020Rule2025} that suggest tokens with high entropy as the most informative signal, we modify the estimator to capture valuable fluctuation instead of highly confident predictions from low-mask sequences.

First, to allocate the timestep budget effectively, we clamp the sampling range from $(0,1)$ to $(\gamma,1)$, where $\gamma$ serves as a cut-off of low mask ratios. Second, rather than relying on randomly masking, we design low-discrepancy estimators that utilize the autoregressive biases in language and locality-driven correlations in vision. We implement the reverse process by managing independent sets of random seeds on each device, which is crucial for stable importance and KL computation. With our designed operator $\text{Rev}(\cdot,t)$, we obtain stable, low-variance estimates driven by the stochasticity of $\hat o_t\sim \text{Rev}(o,t)$.

\noindent\textbf{Let language tokens fade out.} Discrete diffusion models exhibit a causal bias for language~\cite{gongDiffuCoderUnderstandingImproving2025}, particularly in logically related tasks such as math and code. This property, referred to as AR-ness, has been identified to have a strong correlation with model's overall performance. Besides, as the semi-autogressive sampler (Alg.~\ref{algo-semi}) is utilized, the rollouts also exhibit higher divergence as block extends, $i.e.,$ at the start of response, the model's reasoning are rather simple setups, while real divergence or \textit{aha!} moment emerges as the reasoning proceeds. This observation motivates us to exploit the importance estimation through an AR-like reversing procedure, and assign higher attention to the latter tokens. The algorithm shown in Alg.~\ref{algo-armask} maintains a delicate balance between \textit{randomness} and \textit{fading-out} property with almost no additional calculation, and serves as a plug-and-play module.

\begin{minipage}{0.50\textwidth}
\begin{algorithm}[H]
\caption{AR-like Reversing (for text, ours)}
\label{algo-armask}
\begin{algorithmic}[1]
\Require Token batch $X \in \mathbb{R}^{B \times L}$, prompt indicator $C$, mask token $\mathbf{m}$, seed $s$, ratio $r$

\State Set random seed with $s$
\State Prompt (padded) length $L_c \gets \sum C$ 
\State Non-prompt length $L_o \gets L - L_c$ 
\State Linear decay $d \gets \text{linspace}(1,0,L_o)$

\State Normalize $p_n \gets \frac{d \cdot ((1-r) L_o)}{\sum d}$ 
\State $p \gets 0^{L_c} \oplus p_n$,$P \gets \text{repeat}(p,B)$

\State $R \sim U(0,1)^{B \times L}$, $M \gets (\neg C) \land (R > P)$
\State Apply masking $\tilde{X} \gets \text{where}(M, \mathbf{m}, X)$

\State \Return $\tilde{X}, M$
\end{algorithmic}
\end{algorithm}
\end{minipage}
\hfill
\begin{minipage}{0.49\textwidth}
\begin{algorithm}[H]
\caption{Random Reversing (for image)}
\label{algo-randommask}
\begin{algorithmic}[1]
\Require Token batch $X \in \mathbb{R}^{B \times L}$, prompt indicator $C$, mask token $\mathbf{m}$, seed $s$, ratio $r$
\Statex{\ \textcolor{teal}{\texttt{\#\ randomness is managed}}}
\vspace{1pt}
\State Set random seed with $s$
\vspace{1pt}
\Statex{\ \textcolor{teal}{\texttt{\#\ similar as in $q(x_t|x_0,t)$}}}
\vspace{1pt}
\State Constant curve $p \gets r^L$
\State Expand to batch size $P \gets \text{repeat}(p,B)$
\State Sample random matrix $R \sim U(0,1)^{B \times L}$
\State Determine mask $M \gets (\neg C) \land (R < P)$
\State Apply masking $\tilde{X} \gets \text{where}(M, \mathbf{m}, X)$
\vspace{2pt}
\State \Return $\tilde{X}, M$
\end{algorithmic}
\end{algorithm}
\end{minipage}


\subsection{Rollout Adaption}\label{sec:language}
\begin{figure}[htbp]
    \centering
    \includegraphics[width=\linewidth]{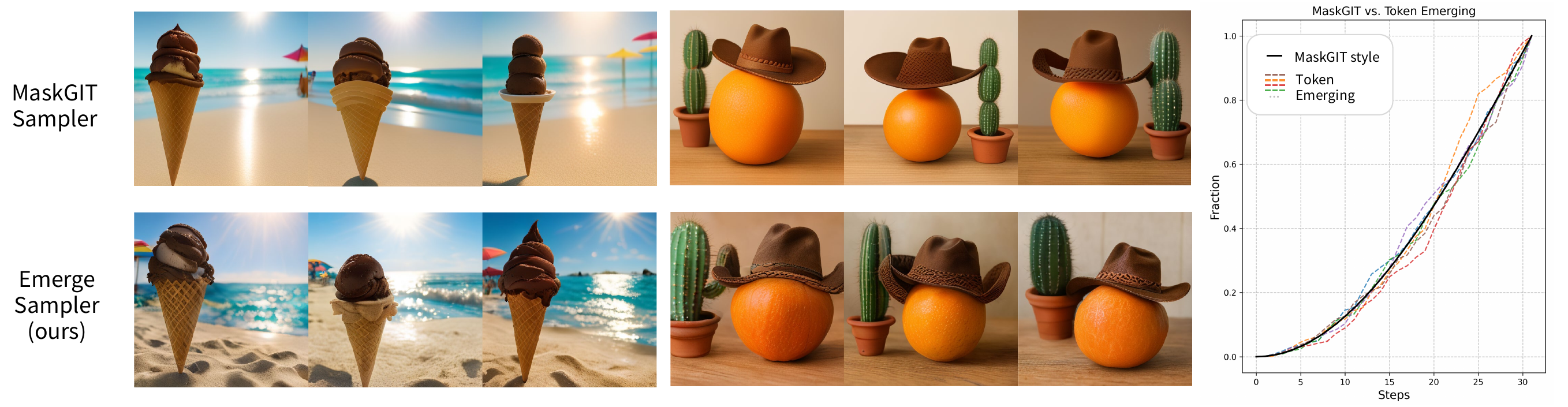}
    \caption{\textbf{A comparison of sampled results.} With identical sampling parameters on MMaDA (equipped with a 8192-vocab visual tokenizer~\cite{xieShowoOneSingle2024}), images sampled by our emerge method (below) demonstrate better texture and expressiveness.}
    \label{fig:example}
\end{figure}
Complementing the estimators, we align the rollout sampler with structural patterns, enabling efficient and stable training with GRPO. 
For text generation, we employ the widely adopted semi-autoregressive sampler (Alg.~\ref{algo-semi}) for optimal generation quality. This sampler uses a low-confidence re-mask strategy with block-wise decoding, reflecting the inherently sequential structure of language.

\noindent\textbf{Let visual tokens emerge from masks.} While expressed as discrete vocabularies, visual tokens differ from language significantly in terms of entropy, bits of information and so on~\cite{chanAnalyzingLanguageVisual2024}. The low-confidence based MaskGIT sampler, which was proposed with a $1024$-vocab tokenizer~\cite{changMaskGITMaskedGenerative2022}, while viable, does not perform as expected for high-fidelity tokenizers. This curse has been partly discussed as sampling inaccuracy~\cite{zhengMaskedDiffusionModels2025}, yet we noticed that it becomes severe on large-vocabulary visual tokenizers and cannot be addressed by simply operating at higher precision. To overcome this problem which hinders model's potential on visual generation, we refer to MDLM~\cite{sahooSimpleEffectiveMasked2024} and propose the probability-based emerging sample strategy. As shown in Fig~\ref{fig:example} and Alg.~\ref{algo-emerge}, our sampler does not enforce a decoding quantity per prediction, but let the visual tokens emerge from masks naturally with probabilistic control. While faithful to the principled DDM theory (Eq.~\ref{reverse}), our sampler shows significantly better expressiveness for vision.

As for reversing, unlike continuous diffusion RL methods~\cite{liuFlowGRPOTrainingFlow2025} where the importance is calculated across almost all traversed timesteps, we find that discrete visual diffusion requires a large truncation on reverse range.The tokenized patches show strong global correlations, making the prediction largely insensitive to small mask ratios (see Sec.\ref{sec:maskrate} for details), and small truncation may even lead to exploded variance. 
Therefore, we keep the reversing random as Alg.~\ref{algo-randommask}, while the reverse strength is held at high level, \textit{e.g.,} setting $\gamma=0.8$ to obtain meaningful importance estimates.

\begin{minipage}{0.53\textwidth}
\begin{algorithm}[H]
\caption{Semi-autoregressive (text)}
\label{algo-semi}
\begin{algorithmic}[1]
\State \textbf{Inputs:} prompt $c$, completion length $L$ 
\State \textbf{Settings:} block length $L_{\text{block}}$, token per step $s$
\State Initialize: $x \gets \mathbf{m}^L$.
\For{$b = 1$ to $L/L_{block}$}
    \State $\text{range}\gets [(b-1)L_{\text{block}}, bL_{\text{block}}]$
    \State $x_b \gets x[\text{range}]$
    \For{$k = 1$ to $L_{\text{block}}/s$}
        \State $p_{\text{conf}} \gets f_\theta(x, c)[\text{range}]$
        \State $x_b \gets \text{where}(x_b=\textbf{m}, \text{argmax}(p_{\text{conf}}), x_b)$
        \State $m_\text{re} \gets \text{argsort}(p_{\text{score}})[: L_{\text{block}}-ks]$
        \State $x_b \gets \text{where}(m_\text{re}, \textbf{m}, x_b))$
    \EndFor
    \State $x[\text{range}]\gets x_b$
\EndFor
\State \textbf{Return:} fully unmasked sequence $x$
\end{algorithmic}
\end{algorithm}
\end{minipage}
\hfill
\begin{minipage}{0.46\textwidth}
\begin{algorithm}[H]
\caption{Token Emerge (image, ours)}
\label{algo-emerge}
\begin{algorithmic}[1]
\State \textbf{Inputs:} prompt $c$, completion length $L$
\State \textbf{Settings:} total steps $K$, scheduler $\alpha_t$
\State Initialize: $x_1 \gets \mathbf{m}^L$.
\For{$k = 1$ to $K$}
\vspace{3pt}
    \State $t \gets \frac{K - k+1}{K}, s \gets \frac{K - k}{K}$
    \State $\text{logits} \gets f_\theta(x_t, c)$
    \vspace{1pt}
    \Statex{\ \textcolor{teal}{\texttt{\#\ Infer from probability}}}
    \vspace{1pt}
    \State $\pi \gets \text{Softmax}(\text{logits})$
    \State $q_s\gets \frac{\alpha_s - \alpha_t}{1- \alpha_t} \cdot\pi + \delta_{\mathbf{m}}\cdot\frac{1-\alpha_s}{1-\alpha_t}$
    \vspace{1pt}
    \Statex{\ \textcolor{teal}{\texttt{\#\ $\mathbf{m}$ is also sampled}}}
    \vspace{1pt}
    \State $x_{\text{pred}} \gets \hat x\sim\text{Cat}(\hat x;q_s)$
    \State $x_s \gets \text{where}(x_t =\mathbf{m}, x_{\text{pred}}, x_t)$
\vspace{3pt}
\EndFor
\State \textbf{Return:} fully unmasked sequence $x_0$
\end{algorithmic}
\end{algorithm}
\end{minipage}
%
\subsection{Algorithm Framework}
To summarize, we implement GRPO for discrete diffusion models with integrated efficient modality-specific importance estimators, and modification on visual sequence sampling. The overall framework consists of: (\textit{i}) \textcolor{db}{\bf Sample} a set of full completions for each prompt $\mathbf{c}$ with method ~\ref{algo-semi},~\ref{algo-emerge}, (\textit{ii}) for each completion $o$, generate multiple masked $\hat o_t\sim \textcolor{db}{\textbf{Rev}}(o,t)$, where $ \textcolor{db}{\textbf{Rev}}(\cdot,t)$ is our designed reverse function with controllable randomness, as shown in Alg.~\ref{algo-armask},~\ref{algo-randommask}. (\textit{iii}) estimating per-completion advantages $A$ based on reward $r$, and (\textit{iv}) updating the policy using importance $\hat\rho^{t_j}$ and divergence $\hat{\mathbb{D}}_{\text{KL}}^{t_j}$.
An algorithmic demonstration of \grpo is provided in Alg.~\ref{algo-grpo}.
\begin{algorithm}
\caption{MaskGRPO Policy Gradient Optimization (ours)}
\label{algo-grpo}
\begin{algorithmic}[1]
\Require Reference model $\pi_{\text{ref}}$, prompt distribution $\mathcal{D}$, number of completions per prompt $G$, number of inner updates $\mu$, random seed set $\mathcal{S}_{j=1\sim \mu}$, 
\State Initialize policy $\pi_\theta \gets \pi_{\text{ref}}$
\While{not converged}
    \State $\pi_{\text{old}} \gets \pi_\theta$
    \State Sample a prompt $\mathbf{c} \sim \mathcal{D}$
    \State  \textcolor{db}{\bf Sample} $G$ completions $o_i \sim \pi_{\text{old}}(\cdot \mid \mathbf{c}), \; i \in [G]$
    \State For each $o_i$, compute reward $r_i$ and advantage $A_i$ using Eq.~\ref{adv}
    \For{gradient update iterations $j = 1, \dots, \mu$}
        \State Get timestep: $t_j \gets \gamma +(1-\gamma)\frac{j}{\mu}$
        \State Construct masked completion $\hat o_{i,t_j} \sim$ \textcolor{db}{\bf Rev}$(o_i, t_j, \mathcal{S}_j )$ 
        \State For $\pi_\theta, \pi_{\text{old}}, \pi_{\text{ref}}$, use Eq.~\ref{rho-ddm},~\ref{kl-ddm} to estimate importance $\hat\rho_i^{t_j}$ and $\hat{\mathbb{D}}_{\text{KL}}^{i,t_j}$ with $\hat o_{i, t_j}$
        \State Compute MaskGRPO objective in Eq.~\ref{grpo-ddm} and update $\pi_\theta$ via gradient descent
    \EndFor
\EndWhile
\State \Return $\pi_\theta$
\end{algorithmic}
\end{algorithm}
\section{Experiment}

\subsection{Setup}\label{sec:setup}
We conduct experiments across multimodal scenarios and evaluated \grpo extensively on math reasoning, coding, and text-to-image generation benchmarks. For language tasks, following the practice of \textit{diffu}-GRPO, we conduct with a similar learning rate of $3e^{-6}$, rollouts per prompt $G=6$, and a global batch size of $96$ ($\text{bs}=6$ on $8\times$A100 GPU, with gradient accumulation $n=2$). The rollout is sampled with a block length of $16$, and $2$ tokens per step. We use iteration $\mu=6$, and train $6000$ steps on all language tasks, which takes up to 25\% fewer training steps and 50\% less iteration than that of \textit{diffu}-GRPO. 

For image generation, we use rollouts per prompt $G=9$ for exploration, iteration $\mu=8$, and a global batch size of $72$ (we exclude two GPUs for reward servers). Each rollout is sampled with our emerge sampler, using 12 steps with a cosine scheduler. The RL training takes $1500$ global steps. Reward design and evaluation details are provided in Appendix.~\ref{appen-details}. 
\paragraph{Models}
We select LLaDA-8B-Instruct~\cite{nieLargeLanguageDiffusion2025a}, an open-sourced native DDM and its multimodal adaption MMaDA-8B-Base~\cite{yangMMaDAMultimodalLarge2025}, which unlocks the ability to perform discrete diffusion on image sequences, as the start point of optimization. Both models are initialized from publicly available pre-trained checkpoints.

\paragraph{Metrics}
We evaluate the proposed \grpo framework on text and image generation tasks, using a suite of standard benchmarks to assess its performance gain.
(1) \textbf{Text Generation}. To evaluate model's mathematical reasoning and coding capabilities, we use GSM8K~\cite{cobbeTrainingVerifiersSolve2021}, MATH500~\cite{lightmanLetsVerifyStep2023} and MBPP~\cite{austinProgramSynthesisLarge2021} benchmarks. We also compare with the recent RL baselines including \textit{diffu}-GRPO~\cite{zhaoD1ScalingReasoning2025}, \texttt{wd1}~\cite{tangWd1WeightedPolicy2025a} and UniGRPO (re-implemented due to unavailable codebase)~\cite{yangMMaDAMultimodalLarge2025} on these tasks.
(2) \textbf{Image Generation}. To evaluate model's text-image alignment, we first utilize the widely adopted GenEval~\cite{ghoshGENEVALObjectFocusedFramework} and DPG-Bench~\cite{huELLAEquipDiffusion2024} (see Appendix~\ref{append-samples}) as the metrics. Then, we evaluate the generated samples' aesthetic quality using human preference scorers like DeQA~\cite{youTeachingLargeLanguage2025}, ImageReward~\citeyear{xuImageRewardLearningEvaluating}, and HPSv3~\cite{maHPSv3WideSpectrumHuman2025a}. For references rather than definitive comparisons, we include both specialized diffusion models, such as SDXL, and leading discrete generation models like Show-o~\cite{xieShowoOneSingle2024} and Janus-Pro~\cite{chenJanusProUnifiedMultimodal2025}. 

\subsection{Performance and Comparison Results}\label{sec:results}

\begin{figure}[htbp]
    \centering
    \includegraphics[width=\linewidth]{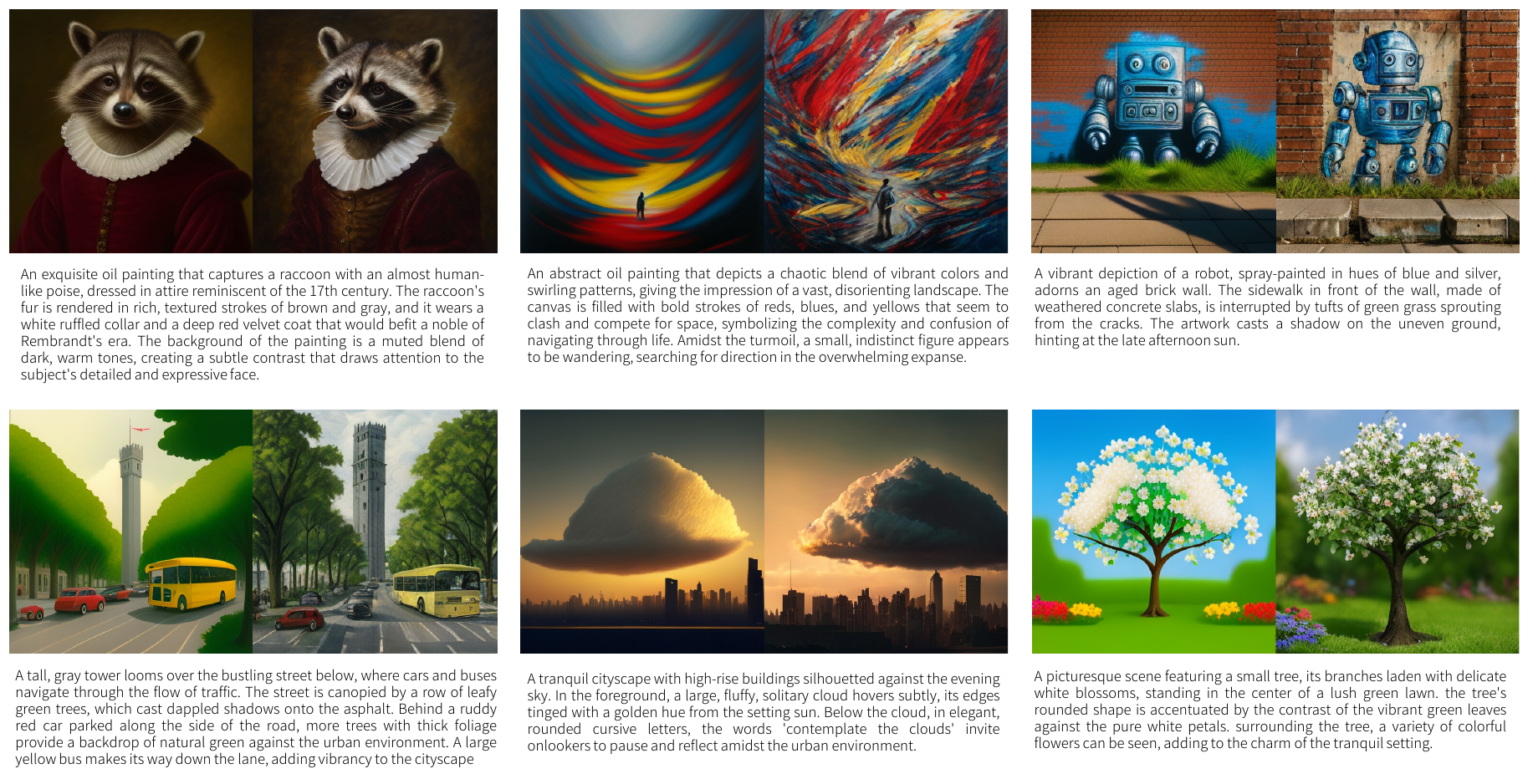}
    \caption{\textbf{Qualitative comparison.} Results are generated with identical sampling parameters and shown in \{\textit{original, w/ RL}\} pairs. \grpo demonstrates substantial improvement on the aesthetic quality of generated images, in terms of artistic style, photographic details and overall atmosphere. We strongly recommend that the readers view more portrait samples at Fig.~\ref{fig:portraits}.}
    \label{fig:qualitative}
\end{figure}

\paragraph{Language Tasks}
\grpo substantially enhances the mathematical reasoning and coding capabilities of LLaDA. As shown in Table~\ref{table:math}, our method achieves over 5\% absolute improvement in solution accuracy on GSM8K, MATH500, and MBPP, nearly doubling the RL gains compared to prior methods with less steps (6000 vs 7000+).
On GSM8K, \grpo allows the model to surpass previous approaches while requiring only half the completion length (256 vs 512), demonstrating its effective improvement of reasoning ability. Representative examples are included in Appendix~\ref{append-samples}.
\begin{table}[t]
\caption{\textbf{Evaluation on math reasoning and coding benchmarks.} For fair comparison, we choose LLaDA-8B-Instruct as the initial point. All results are reported with zero-shot prompting and pass@1 metric. $^\dagger$ refers to our re-implementation. }
\label{table:math}
\centering
\resizebox{0.95\textwidth}{!}{%
\begin{tabular}{lcccccc}
\toprule
\multirow{2}{*}{\textbf{RL Method / Seq Len}} & \multicolumn{2}{c}{\textbf{GSM8K}} & \multicolumn{2}{c}{\textbf{MATH500}} & \multicolumn{1}{c}{\textbf{MBPP}} \\
\cmidrule(lr){2-3} \cmidrule(lr){4-5} \cmidrule(lr){6-6}
 & \textbf{256} & \textbf{512} & \textbf{256} & \textbf{512} & \textbf{256} \\
\midrule
LLaDA-8B-Instruct &76.7 & 78.2 & 32.4 & 36.2 & 39.0  \\
\midrule
\ w/ \textit{diffu}-GRPO~\cite{zhaoD1ScalingReasoning2025} & 79.8 (+3.1)& 81.9 (+3.7)& 34.4 (+2.0) & 39.0 (+2.8) & 42.1 (+3.1)   \\
\ w/ UniGRPO$^\dagger$~\cite{yangMMaDAMultimodalLarge2025} & 81.1 (+4.4) & 82.0 (+3.8) &35.0 (+2.6) & 38.8 (+2.6) & 43.1 (+4.1)  \\
\ w/ \texttt{wd1}~\cite{tangWd1WeightedPolicy2025a} & 80.8 (+4.1)& 82.3 (+4.1)& 34.4 (+2.0) & 39.0 (+2.8) & --   \\
\ w/ TraceRL$^\dagger$~\cite{wangRevolutionizingReinforcementLearning2025} & 82.1 (+5.4)& 83.3 (+5.1)& 35.9 (+3.5) & 39.5 (+3.3) & 43.9(+4.9)   \\
\ w/ \grpo(ours) & \textbf{84.2} \color{db}\bf (+7.5) & \textbf{85.3} \color{db}\bf (+7.1) & \textbf{37.6} \color{db}\bf (+5.2) & \textbf{41.5} \color{db}\bf (+5.2) & \textbf{45.4} \color{db}\bf (+6.4)  \\
\bottomrule
\end{tabular}%
}
\end{table}
\paragraph{Visual Generation Tasks}
To our knowledge, \grpo is the first method to achieve effective GRPO optimization of aesthetic quality and text–image alignment in discrete diffusion models. Table~\ref{complex} shows consistent improvements in alignment with human preferences, which are not reported in previous DDM works.
Moreover, results on GenEval (Tab.~\ref{tab:geneval}) and DPG-Bench (Tab.~\ref{tab:dpg}) further confirm the effectiveness of our framework: with a well-designed RL setup, discrete generation models can approach the performance of leading commercial systems. Qualitative samples are provided in Fig.~\ref{fig:qualitative}. 

\begin{table}[htbp]
\centering
\caption{\textbf{Evaluation on GenEval.} SFT indicates that we SFT the base model with BLIP3-o dataset~\cite{chenBLIP3oFamilyFully2025} for clean instruction-tuning data distilled from GPT-4o.}
\label{tab:geneval}
\resizebox{0.95\textwidth}{!}{%
\begin{tabular}{lccccccc}
\toprule
\multirow{2}{*}{\textbf{Model}}  & \multicolumn{7}{c}{\textbf{GenEval}$\uparrow$} \\
\cmidrule(lr){2-8}
   & \textbf{Single.} & \textbf{Two.} & \textbf{Count.} & \textbf{Color.} & \textbf{Pos.} & \textbf{Attr.} & \textbf{Overall} \\
\midrule
\multicolumn{8}{l}{\textit{Continuous Generation}} \\
SDXL~\cite{podellSDXLImprovingLatent2023}  & 0.98 & 0.74 & 0.39 & 0.85 & 0.15 & 0.23 & 0.55 \\
DALL-E 3~\cite{dalle3betkerImprovingImageGeneration}  & 0.96 & 0.87 & 0.47 & 0.83 & 0.43 & 0.45 & 0.67 \\
SD3.5-L~\cite{esserScalingRectifiedFlow2024}  & 0.98 & 0.89 & 0.73 & 0.83 & 0.34 & 0.47 & 0.71 \\
FLUX.1-dev~\cite{BlackforestlabsFlux2025}  & 0.98 & 0.93 & 0.75 & 0.93 & 0.68 & 0.65 & 0.82 \\
\midrule
\multicolumn{8}{l}{\textit{Discrete Generation}} \\
Show-o~\cite{xieShowoOneSingle2024}  & 0.95 & 0.52 & 0.49 & 0.82 & 0.11 & 0.28 & 0.53 \\
Janus-Pro~\cite{chenJanusProUnifiedMultimodal2025}  & 0.99 & 0.89 & 0.59 & 0.90 & 0.79 & 0.66 & 0.80 \\
MMaDA~\cite{yangMMaDAMultimodalLarge2025}  & 0.96 & 0.60 & 0.45 & 0.81 & 0.14 & 0.25 & 0.56 \\
\ w/ UniGRPO\cite{yangMMaDAMultimodalLarge2025}  & 0.99 & 0.76 & 0.61 & 0.84 & 0.20 & 0.37 & 0.63 \\
\ w/ \grpo(ours)  & 0.99 & 0.85 & 0.66 & 0.89 & 0.73 & 0.69 & 0.80 \\
\ w/ SFT+\grpo(ours)  & 0.99 & 0.85 & 0.69 & 0.90 & 0.77 & 0.72 & 0.84 \\
\bottomrule
\end{tabular}%
}
\end{table}

\begin{table}[h!]
\centering
\caption{\textbf{Evaluation on compositional generation and human preference metrics.} We calculate the Preference Scores on samples generated by DPG-Bench prompts.}
\label{complex}
\resizebox{0.95\textwidth}{!}{%
\begin{tabular}{lccccc}
\toprule
\bf Model & \multicolumn{2}{c}{
\bf Compositional Generation} & \multicolumn{3}{c}{\bf Preference Scores} \\
\cmidrule(lr){2-3} \cmidrule(lr){4-6}
& \bf GenEval & \bf DPG-Bench & \bf DeQA & \bf ImageReward & \bf HPSv3 \\
\midrule
MMaDA~\cite{yangMMaDAMultimodalLarge2025} & 0.56 & 0.71 & 3.99 & 0.93 & 8.81   \\
\midrule
w/ MaskGRPO & 0.80 & 0.75 & 4.10 & 1.18 & 9.40   \\
w/ SFT+MaskGRPO & 0.84 & 0.82 & 4.18 & 1.30 & 9.63   \\
\bottomrule
\end{tabular}%
}
\end{table}

\subsection{Discussion}\label{sec:ablation}
\begin{figure}[htbp]
    \centering
    \includegraphics[width=\linewidth]{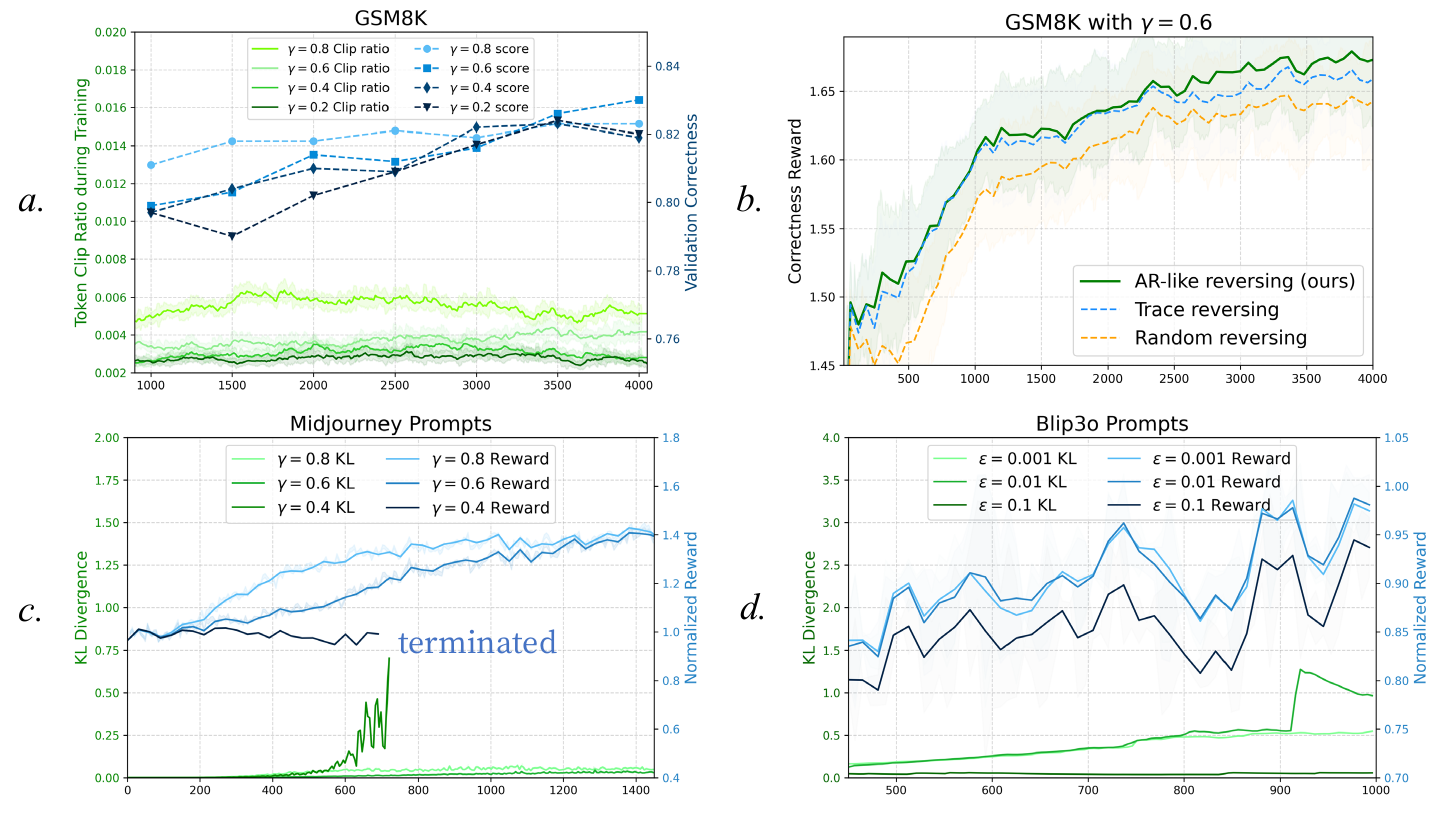}
    \caption{\textbf{Figures for ablative studies.} \textit{a}: ablation on timestep truncation in language tasks. \textit{b}: ablation on reverse methods in language tasks. \textit{c}: ablation on timestep truncation in vision tasks. \textit{d}: ablation for clip range in vision tasks. See text for detailed explanation.}
    \label{fig:ab_curve}
\end{figure}

\paragraph{Truncation hyper-parameters}\label{sec:maskrate}
We perform ablation studies on GSM8K with timestep truncation ratios $\gamma \in {0.2, 0.4, 0.6, 0.8}$ for 4000 steps. As shown in Fig.~\ref{fig:ab_curve} (\textit{a}), both the absence of truncation and overly aggressive truncation degrade training stability. To promote stable learning rather than premature convergence, we adopt $\gamma=0.6$ as the default setting.

\begin{figure}[htbp]
    \centering
    \includegraphics[width=\linewidth]{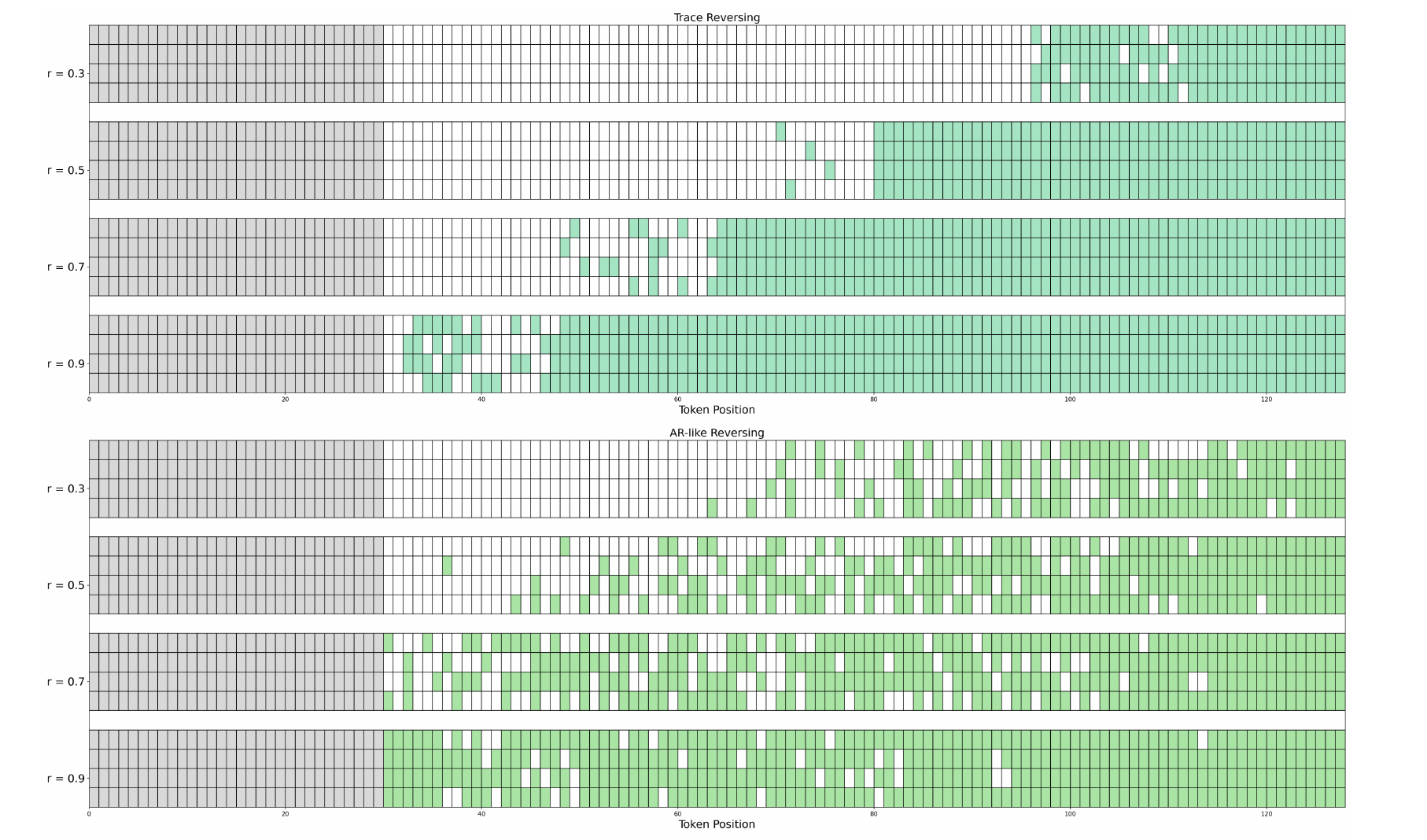}
    \caption{\textbf{A comparison of reversing methods.} Language are decoded in a semi-autoregressive manner, and the trace method~\cite{wangRevolutionizingReinforcementLearning2025} (above) reverses the decoding path accordingly. AR-like methods (below, ours), in contrast, balance the autoregressive bias and randomness with controlled probability, and can capture front fluctuations in the sequence at smaller masking ratios.}
    \label{fig:reverse_with_trace}
\end{figure}

For vision tasks, we conduct experiments with HPSv3 as the primary reward model in (\textit{c}). At $\gamma=0.4$, the strong correlation among tokenized image patches leads to boosting divergence between old and new policies, causing training failure. Based on this, we set $\gamma=0.8$ for efficient and stable training in image generation. Besides, we also conduct experiments on the clip parameter $\epsilon$. As shown in (\textit{d}), relatively small $\epsilon$ ensures better training dynamics for visual tasks, unlike language tasks where $\epsilon$ is usually set to $0.2$ or larger. We attribute this to the global correlation and higher entropy of visual token prediction, a core reason that we adapt modality-specific design for reinforcing DDMs.
\paragraph{Reverse method}\label{sec:tracerl}
Concurrent work TraceRL~\cite{wangRevolutionizingReinforcementLearning2025} proposes to track the generation trace of DDMs and reverses strictly along these recorded traces. This mechanism relies on predefined paths and has only been demonstrated on block-attention architecture (SDAR~\cite{JetAstraSDAR2025}). Moreover, TraceRL requires maintaining trace maps throughout training, and its deterministic reversal leads to limited flexibility in estimating prior tokens. We re-implement TraceRL on the full-attention language model, namely LLaDA-8B-Instruct, and report results in Tab.~\ref{table:math}. In addition, we fix $\gamma = 0.6$ and ablate the reverse strategies in Fig.~\ref{fig:ab_curve} (\textit{b}). Our proposed AR-like reversing method consistently outperforms TraceRL in reinforcement learning. We attribute this performance gap to TraceRL’s path-dependent formulation, which constrains exploration and induces biased estimation of sequence-level importance. For further qualitative evidence, we provide a visualized comparison of reversing strategies under varying ratios in Fig.~\ref{fig:reverse_with_trace}.
%
\paragraph{Rollout comparison}\label{sec:imagerl}
We investigate how the proposed emerge sampler improves generation quality under reinforcement learning. As an initial step, we substitute the vanilla MaskGIT style sampler with our method and evaluate performance on MMaDA using GenEval. Before RL, our method produces samples with better textures, but the GenEval score is worse than that of vanilla method (0.51 vs. 0.56). This discrepancy arises because some of our outputs sometimes exhibit unstable or deformed object boundaries, which negatively affect detector-based metrics.

After RL training, however, these instabilities are largely eliminated. The emerge sampler not only facilitates broader exploration during policy optimization, but also guides the model toward higher-quality local optima that would otherwise be inaccessible to MaskGIT sampling. Consequently, our method achieves a higher GenEval score (0.84 vs. 0.77), while also producing more stable and expressive generations. This progression highlights a key advantage: although the emerge sampler may underperform at the pre-RL stage, its enhanced exploration dynamics ultimately yield stronger convergence and superior sample quality compared to the vanilla baseline. 

\section{Related work}\label{sec:bg}

\paragraph{Text Diffusion Models}
Discrete diffusion models have emerged as a powerful alternative~\cite{sahooSimpleEffectiveMasked2024,nieLargeLanguageDiffusion2025a} to autoregressive paradigms for language generation. Recent works~\cite{zhuLLaDA15VarianceReduced2025, gongDiffuCoderUnderstandingImproving2025} have shown competitive performance in math or code tasks. To obtain optimal results, while current scaled DDMs are trained on fully random noised corpses, state-of-the-art performances are usually obtained via semi-autoregressive decoding~\cite{arriolaBlockDiffusionInterpolating2025,nieLargeLanguageDiffusion2025a}. While this \textit{inconsistency} leaves space for more sophisticated training design, it also partially demonstrates the causal nature of language modality. 

\paragraph{Discrete Visual Diffusion Models}

Discrete diffusion, or its core idea of predicting on multiple discrete targets, has been applied to visual generation with prior endeavors like MaskGIT~\cite{changMaskGITMaskedGenerative2022}, where a low-confidence re-mask sampler with Gumbel noise is used for inference to enhance sample quality. However, this can limit output diversity and makes reliable likelihood estimation challenging~\cite{zhengMaskedDiffusionModels2025, maReDDiTRehashingNoise2025a}, creating a bottleneck for online RL.

\paragraph{Group Relative Policy Optimization}
GRPO and related reward-based optimization techniques have shown success in reinforcement learning for text generation and reasoning in autoregressive LLMs~\cite{shaoDeepSeekMathPushingLimits2024}. Continuous flow-matching methods also adapt this paradigm in recent works via designed SDE solver~\cite{liuFlowGRPOTrainingFlow2025}. However, GRPO's application is fundamentally limited in discrete diffusion models, as it renders intractability on computing importance sampling weight, due to the lack of a factorized likelihood~\cite{zhuLLaDA15VarianceReduced2025}.

\section{Conclusion}

In this work, we introduce \grpo, a modality-aware extension of Group Relative Policy Optimization for discrete diffusion models. Recalling rollout sampling and likelihood estimation, we developed tailored strategies for language and vision generation: fading-out masking for text and probabilistic decoding for images. Our experiments demonstrate that these design choices substantially improve reasoning accuracy, text-image alignment, and sample diversity. These results highlight the importance of modality-specific samplers and estimators for effective policy optimization, and pave the way for unified reinforcement learning approaches across multimodal discrete diffusion.

\bibliography{ref}
\bibliographystyle{ICLR_2026_style/iclr2026_conference}

\appendix

\section{Discussing the Approximation}\label{sec:proof}

\subsection{ELBO estimation}
The intractable log-likelihood $\log \pi_\theta(y|x)$ in DDMs is often approximated from its evidenced lower bound: \begin{equation}\label{elbo-appen}
\mathcal{B}_\pi(y|x)\triangleq\mathbb{E}_{t\sim\mathcal{U}(0,1)}\mathbb{E}_{y^t\sim q(y^t|y,t)}\ell_{\pi}(y^t,t,y|x) \leq \log \pi(y|x).
\end{equation}
And LLaDA-1.5 provided an estimation with proved low variance as 
\begin{equation}\label{vrpo-appen}
\hat{\mathcal{B}}_\pi(y|x)=\frac{1}{\nu}\sum_{j=1}^{\nu}\frac{1}{t_j}\sum_{k=1}^{|y|}\delta(y_{k}^{t_j},\mathbf{m})\log \pi_\theta(y^{t_j}_k,y|x) \approx \log \pi(y|x),
\end{equation}
where $t_j = j/\nu$ is a discretization of timeline $t$, and $y^{t_j}$ is sampled as $y^{t_j}\sim q(y^{t_j};y,t_j)$.
In practice, the time-weighted term is absorbed in to $\ell_\pi$ as an average on masked tokens, and the simplified form is written as:
\begin{equation}\label{vrpo-appen-simp}
\hat{\mathcal{B}}_\pi(y|x)=\frac{1}{\nu}\sum_{j=1}^{\nu}\ell_{\pi}(y^{t_j},y|x),
\end{equation}
\subsection{Importance estimation }
We discuss the importance estimation in Eq.~\ref{approx}. Note that we do not intend to establish a strict proof for this, but kindly discuss the viable implementation which is shared among current RL works. First, let $o$'s subscript index $k$ temporarily represent its $k$-th element, we recall the loss expression as 
\begin{equation}\label{ell-appen}
    \ell_{\pi}(o^t,o|\mathbf{c})\triangleq\sum_{k=1}^{|o|}\delta(o_{k}^t,\mathbf{m})\log \pi(o_{k}|o^t,\mathbf{c}).
\end{equation}
Given partially masked completion $o^t$, the above term describes deviation between model's prediction $\pi$ on $o^t$'s masked positions. We also use $\dot{o}^t = o^t - o^{t+\delta t}$ to denote the tokens that are unmasked at timestep $t$, with $\delta t=\frac{|\dot{o}^t|}{|o|}$.
For clarity, let $x=(c,o^{t+\delta t})$, $y = \dot{o}^{t}$.
Using the low-variance estimation in Eq.~\ref{vrpo-appen-simp}, we discuss Eq.~\ref{approx} as follows:
\begin{eqnarray}\label{proof-1}
\log\pi_{1}(y|x)-\log\pi_{2}(y|x) &\approx& \hat{\mathcal{B}}_{\pi_1}(y|x) - \hat{\mathcal{B}}_{\pi_2}(y|x) \nonumber\\
&=& \frac{1}{\nu}\sum_{j=1}^{\nu}\Big[\ell_{\pi_{1}}(y^{t_j},y|x)-\ell_{\pi_{2}}(y^{t_j},y|x)\Big]
\end{eqnarray}
Considering the small incremental of $y$ comparing to $x$, and the computational resource allocated to the inner-loop $\nu$ is limited, we can make further approximation by calculating on step $t_j=1$, where all tokens in $\dot{o}^t$ are pending: 
\begin{eqnarray}\label{proof-2}
\frac{1}{\nu}\sum_{j=1}^{\nu}\Big[\ell_{\pi_{1}}(y^{t_j},y|x)-\ell_{\pi_{2}}(y^{t_j},y|x)\Big] &\approx&  \ell_{\pi_{1}}(\mathbf{m}^{|y|},y|x)-\ell_{\pi_{2}} (\mathbf{m}^{|y|},y|x) \nonumber \\
&=& \ell_{\pi_{1}}(\dot{o}^t,o|\mathbf{c},o^{t+\delta t})-\ell_{\pi_{2}}(\dot{o}^{t}, o|\mathbf{c},o^{t+\delta t}) \nonumber \\
&=&  \ell_{\pi_{1}}(\dot{o}^t+o^{t+\delta t},o|\mathbf{c})-\ell_{\pi_{2}}(\dot{o}^{t}+o^{t+\delta t}, o|\mathbf{c}) \nonumber \\
&=&  \ell_{\pi_{1}}(o^t,o|\mathbf{c})-\ell_{\pi_{2}}(o^{t}, o|\mathbf{c})
\end{eqnarray}
\hfill$\blacksquare$

\section{Implementation Details}\label{appen-details}

\subsection{Reward function}
We follow common practices and use a reward system for reinforcement learning. For language tasks, we utilize a simple composed function of formatting and correctness. For image generation, following recent RL works~\cite{gengXOmniReinforcementLearning2025}, the reward is composed with Unified-Reward~\cite{wangUnifiedRewardModel2025}, for scoring text-image alignment, HPSv3~\cite{maHPSv3WideSpectrumHuman2025a}, for assessing the aesthetic quality of the image alongside its alignment, and the classic CLIP Score~\cite{hesselCLIPScoreReferencefreeEvaluation2022}. 
\paragraph{Text generation.}For GSM8K and MATH500, the reward consists of two components:
\begin{itemize}
\item[$\bullet$] Correctness reward: returns 2 for an extracted and correct final answer, and 0 otherwise. 
\item[$\bullet$] Format reward: returns 0.5 if the reasoning process is properly enclosed in \texttt{<reasoning>*</reasoning>}, and 0 otherwise.
\end{itemize}

For MBPP, we adopt DiffuCoder’s scheme, combining correctness and format rewards. The format reward ensures completions are wrapped in $'\ '\ '\ *\ '\ '\ '$, while the correctness reward tests generated code against predefined test cases.

\paragraph{Image generation.}For MaskGRPO on image generation tasks, the reward is composed with:
\begin{itemize}
\item[$\bullet$] \textbf{UnifiedReward}, evaluates image–prompt alignment, divided by $5$ to $[0,1]$.
\item[$\bullet$] \textbf{HPSv3}, assess visual quality and text–image alignment, the score is divided by $5$ to an approximate range $[0,2]$.
\item[$\bullet$] \textbf{CLIP Score}, measures similarity between encoded text and image features, ranging from $[0,1]$ (typically $0.2$–$0.4$). We retain this to stabilize training and mitigate reward hacking.
\end{itemize}
The final image reward is computed as the sum of these three components.

\subsection{Evaluation}
For language tasks, we evaluate all tasks with 0-shot prompting. We use a block length of $16$ and decodes $2$ tokens per step for math tasks, and the MBPP protocol is specified in the following paragraph. All performances are reported using the pass@1 metric. For image generation, the sampler decodes an visual sequence of $1024$ tokens (which represents an image with resolution $512\times512$) with $32$ steps, and is equipped with classifier-free guidance at $3.5$, consistent with the original MMaDA configuration.
\paragraph{MBPP Evaluation Protocol.}
We specify the standardized protocol used for evaluating models on the Mostly Basic Python Problems (MBPP) benchmark~\cite{austinProgramSynthesisLarge2021}. We clarify this protocol to address the significant variance in results reported in the literature, which stems from inconsistent settings for generation parameters \texttt{(gen\_len/step/block\_len)}, different prompt designs, and distinct data subsets (e.g., \texttt{sanitized-mbpp.json}). Such variations impede direct model comparisons.

Following the evaluation setup of LLaDA-8B-Instruct, we specify our standard as follows: The test set consists of the first 500 samples (1-500) from the \texttt{mbpp.jsonl} file in the official dataset. The evaluation is conducted in a zero-shot setting, using the same prompt format as the \texttt{lm-eval} library. The generation parameters are fixed at \texttt{gen\_len/step/block\_len = 256/256/32}. Performance is reported using the pass@1 metric, which measures the percentage of test cases passed on the first attempt.
\subsection{Data usage}
\paragraph{Language tasks.}We use the standard training sets for GSM8K and MATH500. For MBPP, we follow DiffuCoder and use Acecode-87K, an open-source code dataset. Notably, \textit{diffu}-GRPO reported using KodCodeLight-RL-10K, which refers to multiple traverse over the dataset, given its reported $7500$ steps. Besides, as the corresponding implementation is missing from its codebase, we re-implement it and discovered limited effectiveness. We report its best performance on MBPP with our re-implementation on Acecode-87K.

\paragraph{Image generation tasks.}For general prompt following ability, we follow X-Omni by randomly sampling 90K prompts from \texttt{midjourney-prompts}, a dataset of real user instructions, and augment it with 60K compositional GenEval-style prompts from Blip3-o (guaranteed that there is no overlap with the benchmarks). This yields roughly 150K samples, on which we train for one epoch.

Additionally, we utilize instruction tuning data from Blip3-o (which is distilled from GPT-4o) for SFT. This procedure is optional, and we have denoted the corresponding results with explicit SFT mark in the tables. As observed, MMaDA's prompt following ability can be improved with such extremely clean supervision signals, and the generated images have a more accurate demonstration of spatial relationship, and clearer boundary among objects, compared to soley RL results. We train on this dataset for $\sim 1000$ steps with global batch size $128$, with a learning rate of $3e^{-6}$.
\subsection{Clarification}
We select LLaDA-8B-Instruct for language tasks and MMaDA-8B-Base for image generation tasks. Both models share a similar architecture and are initialized from LLaDA-8B-Base. While this choice does not affect our claims on multimodal reinforcement learning, we clarify our rationale: although MMaDA released a MixCoT checkpoint, its performance on math and coding tasks is severely limited. With reasoning enabled, it achieves only 48\% accuracy on GSM8K (zero-shot prompting), about 30\% lower than LLaDA-8B-Instruct with same sampling parameters. We attribute this to potentially insufficient training or a suboptimal recipe leading to catastrophic forgetting. Although MaskGRPO applied to MMaDA-8B-MixCoT yields an improvement of over 6\%, the results are not comparable since prior works consistently use LLaDA as the baseline. For this reason, we do not pursue language experiments with MMaDA, and look forward to a truly unified multimodal DDM foundation without performance compromise.

\section{More Results}\label{append-samples}

Detailed results on DPG-Bench is shown in Fig~\ref{tab:dpg}.
We also provide more generated portraits, math reasoning and coding examples as follows.

\begin{table}[htbp]
\centering
\caption{\textbf{Evaluation on DPG-Bench.} \textit{SFT} indicates that we tune the base model with BLIP3-o dataset (distilled from GPT-4o) for clean supervision data before RL training.}
\label{tab:dpg}
\resizebox{0.92\textwidth}{!}{%
\begin{tabular}{lcccccc}
\toprule
\multirow{2}{*}{\textbf{Model}}  & \multicolumn{6}{c}{\textbf{DPG-Bench}$\uparrow$} \\
\cmidrule(lr){2-7}
   & \textbf{Global} & \textbf{Entity} & \textbf{Attribute} & \textbf{Relation} & \textbf{Other} & \textbf{Overall} \\
\midrule
SDXL~\cite{podellSDXLImprovingLatent2023} & 83.27 & 82.43 & 80.91 & 86.76 & 80.41 & 74.65 \\
Emu3-Gen~\cite{wangEmu3NextTokenPrediction2024} & 85.21 & 86.68 & 86.84 & 90.22 & 83.15 & 80.60 \\
MMaDA~\cite{yangMMaDAMultimodalLarge2025} & 77.52 & 77.52 & 77.67 & 81.56 & 63.13 & 70.51 \\
\ w/ \grpo (ours) & 80.92 & 79.52 & 85.41 & 83.05 & 69.12 & 75.81 \\
\ w/ SFT+\grpo (ours) & 85.96 & 84.05 & 89.76 & 83.43 & 79.88 & 81.76 \\
\bottomrule
\end{tabular}%
}
\end{table}

\begin{figure}[htbp]
    \centering
    \includegraphics[width=\linewidth]{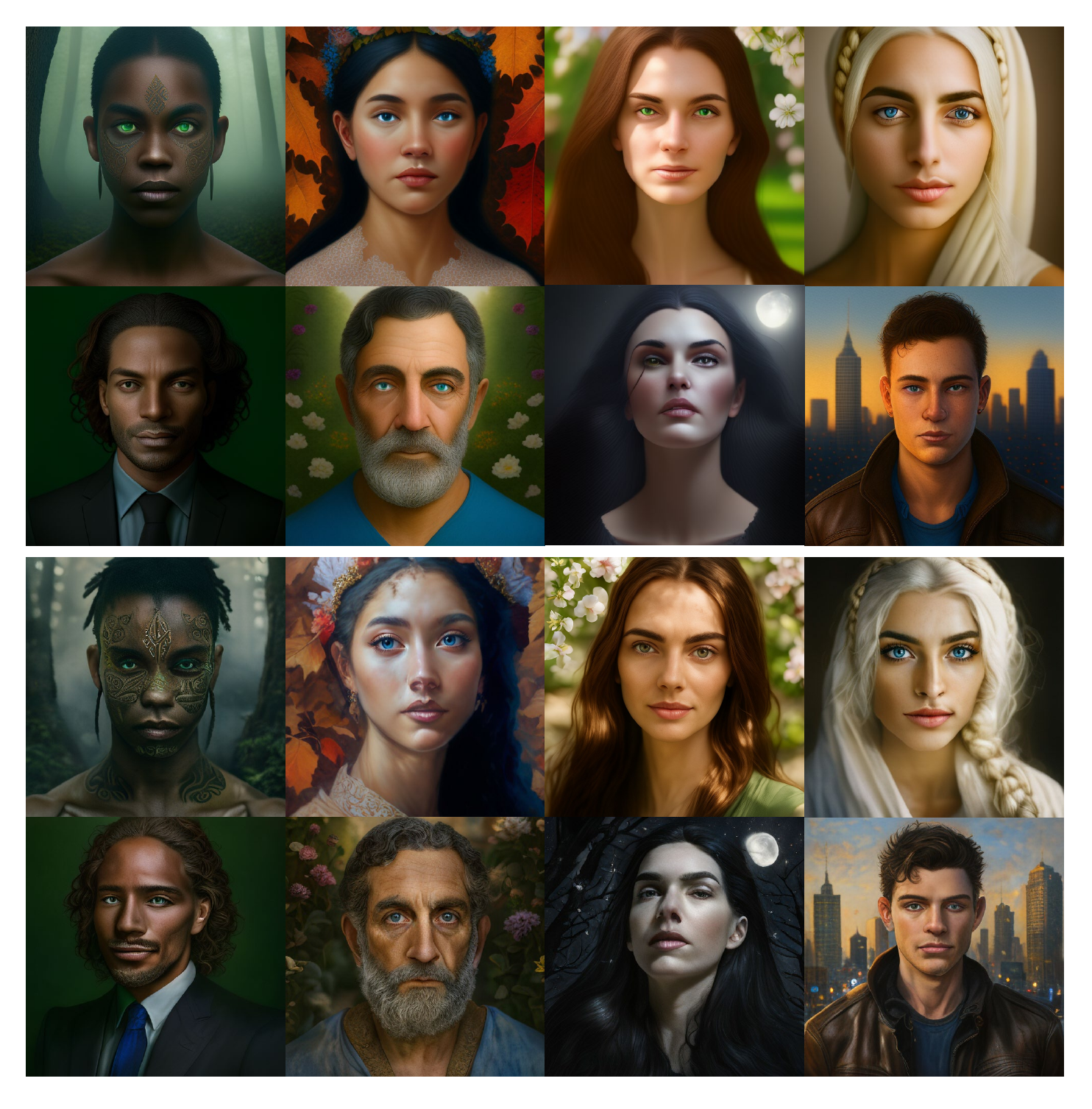}
    \caption{More generated portrait samples with identical prompt and sampling parameters (CFG=3.5, 64 steps). Above: sampled by default MaskGIT-style sampler before RL training. Below: sampled by our emerge sampler after RL training. }
    \label{fig:portraits}
\end{figure}
%

\begin{tcolorbox}[title=\textbf{Generation Sequence Length:} 256 \\ 
\textbf{Question:} There are 29 pupils in a class. The teacher has 9 coupons; each coupon can be redeemed for 100 bottles of apple juice. The teacher gives each student 2 bottles of apple juice to drink for lunch. After redeeming all her coupons and giving each student their apple juice lunch\, how many bottles of apple juice does the teacher have for herself?]
$<$reasoning$>$
First, we need to determine the total number of bottles of apple juice the teacher can redeem with her coupons. Each coupon can be redeemed for 100 bottles, and the teacher has 9 coupons. Therefore, the total number of bottles the teacher can is: 
 \[9 \times 100 = 900\]  
Next, we need to calculate the total number of bottles of apple juice given to the students. There are 29 pupils in the class, and each pupil receives 2 bottles. Therefore, the total number of bottles given to the students is:
 \[29 \times 2 = 58\] 
Now, we subtract the number of bottles given to the students from the total number of bottles the teacher can redeem to find out how many bottles the teacher has for herself:
 \[900 - 58 = 842\] 
Thus, the teacher has 842 bottles of apple juice for herself.
$</$reasoning$>$
\textcolor{green!70!black}{\(\boxed{842}\)}
\end{tcolorbox}

\begin{tcolorbox}[title=\textbf{Generation Sequence Length:} 512 \\ 
\textbf{Question:} A watermelon farm produced 120 baby watermelons. Ashlyn went to the farm after two months and found out that 30\% of the watermelons were ready for harvest\, so she took them home. When she came back two weeks later\, 3/4 of the remaining melons were ready\, so she harvested them. How many melons were not ready to be harvested after the two weeks?]
$<$reasoning$>$
To determine how many watermelons were not ready to be harvested after two weeks, we need to follow these steps:\\
1. Calculate the number of watermelons ready for harvest after two months.\\
2. Subtract the number of watermelons harvested after two months from the total number of watermelons to find the remaining watermelons.\\
3. Calculate the number of watermelons ready for harvest two weeks later.\\
4. Subtract the number of watermelons harvested two weeks later from the remaining watermelons to find the number of watermelons not ready to be harvested.\\
First, let's calculate the number of watermelons ready for harvest after two months:
\[ 30\% \text{ of } 120 = 0.30 \times 120 = 36 \]
So, 36 watermelons were ready for harvest after two months.
Next, we subtract the number of watermelons harvested after two months from the total number of watermelons to find the remaining watermelons:
\[ 120 - 36 = 84 \]
So, there are 84 watermelons remaining after two months.
Now, let's determine how many watermelons were ready for harvest two weeks later:
\[ \frac{3}{4} \text{ of } 84 = \frac{3}{4} \times 84 = 63 \]
So, 63 watermelons were ready for harvest two weeks later.
Finally, we subtract the number of watermelons harvested two weeks later from the remaining watermelons to find the number of watermelons not ready to be harvested:
\[ 84 - 63 = 21 \]
So, 21 watermelons were not ready to be harvested after two weeks.
$</$reasoning$>$
\textcolor{green!70!black}{\(\boxed{21}\)}
\end{tcolorbox}

\begin{tcolorbox}[title=\textbf{Generation Sequence Length:} 256 \\ 
\textbf{Question:} You are an expert Python programmer\, and here is your task: Write a function to find the vertex of a parabola.]
\textbf{Response:}
To find the vertex of a parabola given by the equation 
\[ y = ax^2 + bx + c \]
you can use the following formulas:\\
1. The x-coordinate of the vertex is given by 
\[ x = -\frac{b}{2a} \]
2. The y-coordinate of the vertex is given by 
\[ y = c - \frac{b^2}{4a} \]
Here's a Python function that implements these formulas:
\begin{verbatim}
def parabola_vertex(a, b, c): 
    x_vertex = -b / (2 * a)     
    y_vertex = c - (b**2) / (4 * a)     
    return (x_vertex, y_vertex) 
\end{verbatim}
\textbf{Test cases:} \\
"assert parabola\_vertex(5,3,2)==(-0.3, 1.55)", \textcolor{green!70!black}{\(\boxed{True}\)} \\
"assert parabola\_vertex(9,8,4)==(-0.4444444444444444, 2.2222222222222223)", \textcolor{green!70!black}{\(\boxed{True}\)} \\
"assert parabola\_vertex(2,4,6)==(-1.0, 4.0)", \textcolor{green!70!black}{\(\boxed{True}\)} \\

\end{tcolorbox}

\end{document}